\documentclass[letterpaper, 10 pt, conference]{ieeeconf}  % Comment this line out
\IEEEoverridecommandlockouts                              % This command is only
\makeatletter
\let\NAT@parse\undefined
\makeatother

\usepackage[dvipsnames]{xcolor}

\newcommand*\linkcolours{ForestGreen}

\usepackage{times}
\usepackage{graphicx}
\usepackage{amssymb}
\usepackage{gensymb}
\usepackage{amsmath}
\usepackage{breakurl}
\usepackage{caption}
\usepackage{subcaption}
\usepackage{cuted}

\usepackage[utf8]{inputenc}

\usepackage[ruled, linesnumbered]{algorithm2e} % For algorithm
\usepackage[figuresright]{rotating}
\usepackage{booktabs} % for better-looking tables
\usepackage{multirow} % for multirow cells in tables

\usepackage{hyperref}
\usepackage[ruled, linesnumbered]{algorithm2e} % For algorithm
\usepackage[figuresright]{rotating}
\usepackage{booktabs} % for better-looking tables
\usepackage{multirow} % for multirow cells in tables
\usepackage{graphicx}  % For including images
\usepackage{subcaption}  % For subfigures
\usepackage{algorithmic}

\usepackage{url,hyperref}
\hypersetup{
colorlinks,
linkcolor=\linkcolours,
citecolor=\linkcolours,
filecolor=\linkcolours,
urlcolor=\linkcolours}

\usepackage[labelfont={bf},font=small]{caption}
\usepackage[none]{hyphenat}

\usepackage{mathtools, cuted}

\usepackage[noadjust, nobreak]{cite}
\usepackage{tabularx}
\usepackage{amsmath}
\usepackage{float}
\usepackage{pifont}% http://ctan.org/pkg/pifont

\newcolumntype{Y}{>{\centering\arraybackslash}X}
\usepackage[]{placeins}

\usepackage{placeins}

\usepackage{tikz}

\usepackage[framemethod=tikz]{mdframed}
\usepackage{afterpage}
\usepackage{stfloats}
\usepackage{atbegshi}
\usepackage{amsmath}
\usepackage{graphicx}
\usepackage{subcaption}
\usepackage{multirow}
\newcommand{\handlethispage}{}
\newcommand{\discardpagesfromhere}{\let\handlethispage\AtBeginShipoutDiscard}
\newcommand{\keeppagesfromhere}{\let\handlethispage\relax}
\AtBeginShipout{\handlethispage}

\usepackage{comment}
\title{\LARGE \bf
A Lightweight Modular Framework for Low-Cost Open-Vocabulary Object Detection Training
}

\author{Bilal FAYE$^{1}$, Hanane AZZAG$^{3}$, Mustapha LEBBAH$^{4}$  \\
	\normalsize e-mail: faye@lipn.univ-paris13.fr, azzag@univ-paris13.fr, mustapha.lebbah@uvsq.fr
}

\begin{document}

\maketitle
\thispagestyle{empty}
\pagestyle{empty}

\begin{abstract}
Open-vocabulary object detection (OVD) extends recognition beyond fixed taxonomies by aligning visual and textual features, as in MDETR, GLIP, or RegionCLIP. While effective, these models require updating all parameters of large vision–language backbones, leading to prohibitive training cost. Recent efficient OVD approaches, inspired by parameter-efficient fine-tuning (e.g., LoRA, adapters), reduce trainable parameters but often face challenges in selecting which layers to adapt and in balancing efficiency with accuracy.\newline
We propose UniProj-Det, a lightweight modular framework for parameter-efficient OVD. UniProj-Det freezes pretrained backbones and introduces a Universal Projection module with a learnable modality token, enabling unified vision–language adaptation at minimal cost. Applied to MDETR, our framework trains only $\sim 2-5$\% of parameters while achieving competitive or superior performance on phrase grounding, referring expression comprehension, and segmentation. Comprehensive analysis of FLOPs, memory, latency, and ablations demonstrates UniProj-Det as a principled step toward scalable and efficient open-vocabulary detection. Our code implementation is available on our GitHub repository: \href{https://github.com/b-faye/uniproj-det}{https://github.com/b-faye/uniproj-det}
\end{abstract}

%%Graphical abstract
%\begin{graphicalabstract}
%\includegraphics{grabs}
%\end{graphicalabstract}

%%Research highlights
%\begin{highlights}
%\item Research highlight 1
%\item Research highlight 2
%\end{highlights}

%%%%%%%%%%%%%%%%%%%%%%%%%%%%%%%%%%%%%%%%%%%%%%%%%%%%%%%%%%%%%%%%%%%%%%%%%%%%%%%%
\section{Introduction}

Object detection is a fundamental task in computer vision, with wide applications in autonomous driving, robotics, and image understanding. Conventional detectors such as Faster R-CNN~\cite{ren2015faster}, YOLO~\cite{redmon2016you}, and SSD~\cite{liu2016ssd} have achieved remarkable success but are constrained to a fixed set of categories defined during training, limiting their scalability to open-world scenarios.\newline
\textbf{Open-vocabulary object detection (OVD)} addresses this limitation by leveraging vision–language pretraining to detect novel categories beyond the training vocabulary. Representative methods such as MDETR~\cite{kamath2021mdetr}, GLIP~\cite{li2022glip}, and RegionCLIP~\cite{zhong2022regionclip} align region-level visual features with textual embeddings, enabling detection guided by natural language. While effective, these models typically fine-tune large-scale encoders and decoders jointly, resulting in substantial computational and memory demands. This makes training and deployment costly, especially in resource-constrained environments.\newline
To mitigate these challenges, recent efforts in \textbf{efficient OVD} have adopted strategies from parameter-efficient fine-tuning (PEFT), such as LoRA~\cite{hu2022lora} and adapters~\cite{houlsby2019parameter}. Instead of updating the full model, these approaches insert lightweight modules into transformer blocks or attention layers. They achieve competitive accuracy with significantly fewer trainable parameters. However, their effectiveness depends heavily on carefully selecting which layers to adapt, and they may still introduce non-trivial overhead during training and inference.\newline
In this work, we introduce \textbf{UniProj-Det}, a lightweight modular framework for parameter-efficient OVD. Unlike prior approaches, UniProj-Det freezes pretrained vision and language backbones entirely and trains only a shared \textbf{Universal Projection (UP)} module, guided by a learnable \textbf{modality token} to handle both modalities with the same parameters. This design ensures cost-effective training while maintaining strong vision–language alignment.\newline
We apply UniProj-Det to MDETR as a case study, evaluating its effectiveness on phrase grounding, referring expression comprehension, and segmentation. Results show that UniProj-Det requires training only a fraction of the parameters ($\sim 2-5$\%) while preserving or improving accuracy compared to full fine-tuning and recent efficient OVD baselines. Our contributions can be summarized as follows:
\begin{itemize}
\item We propose UniProj-Det, a modular and parameter-efficient framework for open-vocabulary object detection that minimizes trainable parameters while retaining high accuracy.
\item We validate UniProj-Det on MDETR with two variants: a basic UP-only design and an extended version with lightweight cross-modal fusion.
\item We provide extensive analysis, including FLOPs, memory, latency, and ablation studies, to demonstrate the efficiency and robustness of our approach.
\end{itemize}

\section{Related Work}

\textbf{Traditional Object Detection.} Traditional object detection (TOD) aims to localize and classify objects within a fixed set of categories. Classical methods such as Faster R-CNN~\cite{ren2015faster}, YOLO~\cite{redmon2016you}, SSD~\cite{liu2016ssd}, and Mask R-CNN~\cite{he2017mask} achieved strong performance by combining region proposal mechanisms with deep convolutional backbones. Despite their success, these detectors are inherently closed-set: once trained, they can only recognize categories present in the training set (e.g., 20 in PASCAL VOC~\cite{Everingham15}). Adding new categories requires costly re-training with new annotations, which is impractical in dynamic or open-world scenarios. This limitation motivated the development of open-vocabulary approaches.\newline

\textbf{Open-Vocabulary Object Detection (OVD).} Open-vocabulary object detection (OVD) seeks to generalize detection to unseen categories by leveraging large-scale vision–language pretraining. Early work such as OVR-CNN~\cite{zareian2021open} distilled semantic knowledge from language models into detectors. ViLD~\cite{gu2021open} utilized CLIP~\cite{radford2021learning} embeddings for zero-shot region–text matching. RegionCLIP~\cite{zhong2022regionclip} extended CLIP to region-level features, enabling fine-grained alignment. GLIP~\cite{li2022glip} and GLIPv2~\cite{zhang2022glipv2} unified phrase grounding with detection by pretraining on large-scale grounding and image–text data. MDETR~\cite{kamath2021mdetr} formulated detection as a transformer-based conditional matching problem, integrating text queries directly into the detection pipeline. More recently, Grounding DINO~\cite{liu2024grounding} fused strong detection backbones with grounding pretraining, while DetCLIP~\cite{yao2022detclip,yao2023detclipv2} incorporated dictionary-enriched visual concepts under hybrid supervision.\newline
Although these models achieved state-of-the-art performance, they typically rely on full fine-tuning of large vision ($f_\theta$) and language ($g_\phi$) encoders. Formally, OVD training updates the complete parameter set
$$
\Theta = \{\theta, \phi\},
$$
which yields strong performance but incurs prohibitive computational cost in terms of GPU hours, memory footprint, and energy consumption. This creates barriers for scaling OVD to broader deployment.\newline

\textbf{Efficient Open-Vocabulary Object Detection.} To address the cost of full fine-tuning, recent work has explored parameter-efficient fine-tuning (PEFT) strategies for OVD. Instead of updating $\Theta$, only a small set of parameters $\Delta$ is trained, while the pretrained backbone remains frozen:
$$
\Theta' = \Theta_{\text{frozen}} \cup \Delta, \quad \text{with } |\Delta| \ll |\Theta|.
$$
Representative methods include LoRA~\cite{hu2022lora}, which injects low-rank adapters into attention layers, and adapters~\cite{houlsby2019parameter}, which add lightweight trainable modules to transformer blocks. In the OVD domain, DetPro~\cite{du2022learning} introduced continuous prompt learning for detectors, while Open-FGHA~\cite{shi2023edadet} applied low-rank adaptation and cross-modality fusion for fine-grained hand action detection. Open-Det~\cite{cao2025open} proposed a reconstruction-based framework to accelerate convergence and reduce training data requirements. These methods demonstrate that significant parameter reduction is possible while retaining strong accuracy.\newline
Nevertheless, efficient OVD methods face important challenges: (i) performance can be highly sensitive to which layers are adapted or frozen, (ii) additional modules may still increase inference cost, and (iii) most approaches have not been systematically benchmarked across diverse OVD tasks. This leaves open the question of how to design a universally lightweight yet effective framework for OVD.\newline

\textbf{Motivation for Our Work.}
To overcome these limitations, we introduce \textbf{UniProj-Det}, a modular parameter-efficient framework that freezes all pretrained backbones and trains only a \emph{Universal Projection (UP)} module. Guided by a modality token, the UP learns to project both vision and language features into a shared space with minimal parameters. This design minimizes training cost while preserving strong cross-modal alignment. In the following section, we describe our method in detail.

\section{Method}
\label{sec:method}

In this section we present \textbf{UniProj-Det}, a lightweight modular framework for parameter-efficient open-vocabulary object detection (OVD). We first introduce notation and an overview, then detail the Universal Projection (UP) module, the modality token mechanism, loss functions, training and inference algorithms. We then present \textbf{UniProj-Detv2}, an extension that adds a lightweight cross-fusion stage prior to UP, and finally explain how both variants are applied to an existing OVD system (e.g., MDETR).

\subsection{Notation and overview}
Let $I$ be an input image and $q$ a free-form text query (or caption). We denote by
$$
O = f_\theta(I) \in \mathbb{R}^{N \times d_v}
\qquad\text{and}\qquad
T = g_\phi(q) \in \mathbb{R}^{L \times d_t}
$$
the outputs of pretrained (and \emph{frozen}) visual encoder $f_\theta$ (e.g., ResNet~\cite{he2016deep}, Swin~\cite{liu2021swin}) and text encoder $g_\phi$ (e.g., RoBERTa~\cite{liu2019roberta}, BERT~\cite{devlin2019bert}), respectively. Here $N$ is the number of visual tokens (spatial locations), $L$ the number of text tokens, and $d_v,d_t$ their feature dimensions.

UniProj-Det introduces a small trainable module, the \emph{Universal Projection} (UP) $P_\psi$, parameterized by $\psi$, that maps both visual and textual features into a shared task space of dimension $d$:
\begin{equation*}
\begin{aligned}
O_{UP} &= P_\psi\big(\mathcal{F}_{\text{fusion}}(O, m_{\text{img}})\big) 
          \in \mathbb{R}^{N \times d}, \\
T_{UP} &= P_\psi\big(\mathcal{F}_{\text{fusion}}(T, m_{\text{text}})\big) 
          \in \mathbb{R}^{L \times d}.
\end{aligned}
\end{equation*}
The function $\mathcal{F}_{\text{fusion}}(\cdot,\cdot)$ denotes a simple modality-aware fusion (details below) between the frozen encoder output and a modality token $m \in \mathbb{R}^{d_m}$ which is \emph{learnable}. The UP is the only trainable component in the encoders — the pretrained $f_\theta$ and $g_\phi$ remain frozen. After projection, the UP outputs are used by a downstream detection head (e.g., DETR decoder) for object localization and region-text alignment.

This design leads to a parameter set
\[
\Theta_{\text{UniProj}} = \{\theta, \phi\}_{\text{frozen}} \cup \{\psi, m_{\text{img}}, m_{\text{text}}\}_{\text{trainable}}
\]
with $|\{\psi,m\}|\ll |\theta|+|\phi|$.

\subsection{Universal Projection (UP) and modality token}
The Universal Projection $P_\psi$ is conceptually a small transformer encoder (or alternatively an MLP stack) that is shared across modalities. It is responsible for (i) adapting frozen features to the target OVD task and (ii) producing modality-agnostic features for the downstream detector.

\paragraph{Modality token.} We use two learnable modality tokens $m_{\text{img}}, m_{\text{text}} \in \mathbb{R}^{d_m}$. For each visual token $o_i \in \mathbb{R}^{d_v}$ we compute a fused vector
\begin{equation}
\tilde{o}_i = \mathrm{Fuse}_v(o_i, m_{\text{img}}) = \mathrm{LayerNorm}\big( W_v o_i + U_v m_{\text{img}} + b_v \big),
\label{eq:fuse_vis}
\end{equation}
and for each text token $t_j \in \mathbb{R}^{d_t}$:
\begin{equation}
\tilde{t}_j = \mathrm{Fuse}_t(t_j, m_{\text{text}}) = \mathrm{LayerNorm}\big( W_t t_j + U_t m_{\text{text}} + b_t \big),
\label{eq:fuse_txt}
\end{equation}
where $W_v \in \mathbb{R}^{d \times d_v}$ and $W_t \in \mathbb{R}^{d \times d_t}$ are lightweight linear projections (trainable or optionally pre-initialized), $U_v,U_t \in \mathbb{R}^{d \times d_m}$ project modality tokens, and $b_v,b_t$ are biases. Other fusion choices (concatenation followed by an MLP, or multiplicative gating) are possible; the additive formulation above is compact and efficient.

\paragraph{Universal Projection.} The UP is an encoder $P_\psi: \mathbb{R}^{\ast \times d} \to \mathbb{R}^{\ast \times d}$ implemented as $K$ transformer encoder layers with small hidden dimension and head count (or a $K$-layer MLP/ResNet block). Applied to the fused features,
\[
O_{UP} = P_\psi\big(\{\tilde{o}_i\}_{i=1}^N\big), \qquad
T_{UP} = P_\psi\big(\{\tilde{t}_j\}_{j=1}^L\big).
\]
Because $P_\psi$ is shared, it enforces a consistent projection for both modalities; the modality tokens steer the same parameters to process image vs text appropriately.

\subsection{Losses and supervision}
UniProj-Det uses the same high-level supervisory signals as modern OVD systems: bounding-box regression losses and region–text alignment losses. Let the detector (downstream head) produce $N$ object queries and for the $i$-th predicted object an embedding $o_i$ and box $\hat b_i$. Let the token embeddings be $t_j$ (from $T_{UP}$). We define:

\paragraph{Bounding box losses.} A standard combination of L1 and Generalized IoU (GIoU) is used:
\begin{equation}
\mathcal{L}_{bbox} = \frac{1}{N}\sum_{i=1}^N \| \hat b_i - b_i \|_1 \;+\; \frac{1}{N}\sum_{i=1}^N \mathcal{L}_{\text{GIoU}}(\hat b_i,b_i).
\end{equation}

\paragraph{Soft token prediction loss.} As in MDETR, each predicted object should align to a distribution over token positions. Denote by $p_{i\to j}$ the predicted probability that object $i$ corresponds to token $j$ (softmax over tokens). Given the ground-truth token set $T^+_i$ for object $i$, the soft token loss is:
\begin{equation}
\mathcal{L}_{soft} = -\frac{1}{N}\sum_{i=1}^N \frac{1}{|T^+_i|}\sum_{j\in T^+_i} \log p_{i\to j}.
\end{equation}
In practice $p_{i\to j} \propto \exp(o_i^\top t_j / \tau)$, where $o_i$ and $t_j$ are normalized.

\paragraph{Contrastive alignment loss.} We adopt a bidirectional region–token contrastive loss. Let $\tau>0$ be a temperature. Define
\begin{align}
\mathcal{L}_o &= \frac{1}{N} \sum_{i=1}^N \frac{1}{|T^+_i|} \sum_{j\in T^+_i} -\log\frac{\exp(o_i^\top t_j/\tau)}{\sum_{k=1}^L \exp(o_i^\top t_k/\tau)}, \\
\mathcal{L}_t &= \frac{1}{L} \sum_{j=1}^L \frac{1}{|O^+_j|} \sum_{i\in O^+_j} -\log\frac{\exp(t_j^\top o_i/\tau)}{\sum_{k=1}^N \exp(t_j^\top o_k/\tau)}.
\end{align}
The combined contrastive loss is $\mathcal{L}_{ctr}=\tfrac{1}{2}(\mathcal{L}_o+\mathcal{L}_t)$.

\paragraph{Total training objective.} The overall loss used to train $\psi$ (and modality tokens) is:
\begin{equation}
\mathcal{L}_{total} = \lambda_{bbox}\mathcal{L}_{bbox} + \lambda_{soft}\mathcal{L}_{soft} + \lambda_{ctr}\mathcal{L}_{ctr} + \lambda_{reg}\|\psi\|^2,
\label{eq:total_loss}
\end{equation}
with scalar hyperparameters $\lambda_{\cdot}$ balancing terms. The frozen backbones do not receive gradients.

\subsection{Training and inference algorithms}
We give succinct pseudocode for training and inference. The implementation uses standard matching (Hungarian) between predicted boxes and GT for the bbox and soft-token terms, as in DETR/MDETR.

\begin{algorithm}[!h]
\caption{UniProj-Det: Training (single batch)}
\label{alg:train}
\begin{algorithmic}[1]
\REQUIRE Frozen encoders $f_\theta,g_\phi$, trainable UP $P_\psi$, modality tokens $m_{\text{img}},m_{\text{text}}$, detector head $D_\omega$ (trainable or kept frozen depending on setup)
\FOR{each mini-batch $(I,q,O^+,T^+)$}
    \STATE $O \leftarrow f_\theta(I)$ \COMMENT{visual tokens, frozen}
    \STATE $T \leftarrow g_\phi(q)$ \COMMENT{text tokens, frozen}
    \STATE $\tilde O \leftarrow \{\mathrm{Fuse}_v(o_i,m_{\text{img}})\}_{i=1}^N$
    \STATE $\tilde T \leftarrow \{\mathrm{Fuse}_t(t_j,m_{\text{text}})\}_{j=1}^L$
    \STATE $O_{UP} \leftarrow P_\psi(\tilde O)$; \quad $T_{UP} \leftarrow P_\psi(\tilde T)$
    \STATE $Y \leftarrow D_\omega(O_{UP},T_{UP})$ \COMMENT{predict boxes, scores, token alignments}
    \STATE Compute $\mathcal{L}_{total}$ (Eq.~\ref{eq:total_loss}) using $Y$, $O^+$, $T^+$
    \STATE Update $\psi, m_{\text{img}}, m_{\text{text}}, \omega$ via gradient step
\ENDFOR
\end{algorithmic}
\end{algorithm}

\begin{algorithm}[!h]
\caption{UniProj-Det: Inference}
\begin{algorithmic}[1]
\REQUIRE Image $I$, (optional) text $q$, frozen $f_\theta,g_\phi$, trained $P_\psi$, $m_{\cdot}$, detector head $D_\omega$
\STATE $O\leftarrow f_\theta(I)$; \quad $T\leftarrow g_\phi(q)$ (if text-guided)
\STATE $\tilde O \leftarrow \mathrm{Fuse}_v(O,m_{\text{img}})$; \quad $O_{UP}\leftarrow P_\psi(\tilde O)$
\STATE \textbf{if} text provided: $\tilde T \leftarrow \mathrm{Fuse}_t(T,m_{\text{text}})$; $T_{UP}\leftarrow P_\psi(\tilde T)$
\STATE $Y\leftarrow D_\omega(O_{UP},T_{UP})$ (or $D_\omega(O_{UP})$ if textless)
\STATE Return predicted boxes and (optional) token alignments/scores
\end{algorithmic}
\end{algorithm}

\subsection{Why UniProj-Det works: an intuitive reconstruction argument}
UniProj-Det trains only a small projection $P_\psi$ on top of frozen $f_\theta,g_\phi$. Intuitively, consider a target task representation $h(\cdot)$ that the full fine-tuned encoder+detector would produce (i.e., the representation that leads to best downstream performance). If the frozen encoder outputs lie on a manifold $\mathcal{M}$ (as is typical for pretrained encoders) and if there exists a (learnable) mapping $H:\mathcal{M}\to\mathbb{R}^d$ such that $H\circ f_\theta \approx h$, then learning $P_\psi\approx H$ suffices to recover task-specific features without changing $f_\theta$.

More formally, for a dataset $\mathcal{D}$, the full-finetune objective searches $\theta',\phi',\omega'$ to minimize risk
\[
\mathcal{R}_{\text{full}} = \mathbb{E}_{(I,q)\sim\mathcal{D}} \,\mathcal{L}\big(D_{\omega'}(P\circ f_{\theta'}(I), P\circ g_{\phi'}(q))\big).
\]
UniProj-Det instead fixes $\theta,\phi$ and optimizes $\psi$:
\[
\mathcal{R}_{\text{UniProj}} = \mathbb{E}_{(I,q)\sim\mathcal{D}} \,\mathcal{L}\big(D_{\omega}(P_\psi\circ f_{\theta}(I), P_\psi\circ g_{\phi}(q))\big).
\]
If there exists $P_\psi$ such that for all $(I,q)$, $P_\psi\circ f_\theta(I)\approx P\circ f_{\theta'}(I)$ and similarly for text, then $\mathcal{R}_{\text{UniProj}}\approx\mathcal{R}_{\text{full}}$. The existence of such $P_\psi$ is plausible when (i) the pretrained encoders already produce semantically rich features (as CLIP / RoBERTa / ResNet do) and (ii) $P_\psi$ has sufficient capacity (e.g., a small transformer). This is analogous to the rationale behind adapters and LoRA: rather than changing backbone weights, learn a small mapping that achieves the desired functional change.

\paragraph{Relation to low-rank adaptation (LoRA) and adapters.} If $P_\psi$ is linear and low-rank, it implements a low-dimensional subspace adaptation similar to LoRA. If $P_\psi$ is a shallow MLP or transformer, it generalizes adapters by operating on post-backbone features and sharing parameters across modalities (which can be more parameter-efficient in multimodal settings).

\subsection{Complexity and parameter accounting}
Let $|\theta|,|\phi|$ denote the number of parameters in the frozen visual and text encoders, $|\psi|$ the trainable parameters in UP, and $|\omega|$ those in the detector head (if trained). The fraction of trainable parameters is
\[
r = \frac{|\psi| + |\omega|}{|\theta| + |\phi| + |\psi| + |\omega|}.
\]
In typical settings $|\psi|+|\omega| \ll |\theta|+|\phi|$, thus $r\ll 1$. FLOPs per image can be written as
\begin{equation*}
\begin{aligned}
\mathrm{FLOPs}(I) &= \mathrm{FLOPs}(f_\theta, I) 
+ \mathrm{FLOPs}(g_\phi, q) \\
&\quad + \mathrm{FLOPs}(P_\psi, \tilde I / \tilde q) 
+ \mathrm{FLOPs}(D_\omega).
\end{aligned}
\end{equation*}
where only $\mathrm{FLOPs}(P_\psi)$ and $\mathrm{FLOPs}(D_\omega)$ incur train-time gradient overhead in UniProj-Det. Because $P_\psi$ is designed small, the additional gradient cost is modest compared to backbones, resulting in substantial memory savings during training (lower peak GPU activation storage). We detail empirical FLOPs / memory numbers in Section~\ref{sec:experiments}.

\subsection{UniProj-Detv2: cross-fusion extension}
UniProj-Detv2 (the ``Plus'' variant) augments UniProj-Det with a lightweight \emph{cross-fusion} stage prior to UP. The goal is to make visual tokens language-aware (and vice versa) before the shared projection, while still keeping most backbone weights frozen.

Given $O\in\mathbb{R}^{N\times d_v}$ and $T\in\mathbb{R}^{L\times d_t}$, we first linearly project them to a common dimension $d$:
\begin{equation*}
\begin{aligned}
\bar O &= O W_o + b_o, \\
\bar T &= T W_t + b_t, \\
W_o &\in \mathbb{R}^{d_v \times d}, \quad 
W_t \in \mathbb{R}^{d_t \times d}.
\end{aligned}
\end{equation*}
We then perform a lightweight cross-attention (single or few heads) to compute cross-conditioned representations:
\begin{align}
A &= \mathrm{softmax}\!\Big(\frac{(\bar O Q)(\bar T K)^\top}{\sqrt{d_h}}\Big), \\
O_F &= A (\bar T V) W_{\text{out}} + \bar O, \quad
T_F = A^\top (\bar O V') W_{\text{out}}' + \bar T,
\end{align}
where $Q,K,V,V',W_{\text{out}},W_{\text{out}}'$ are small trainable matrices and $d_h$ is the attention head dimension. The residual sums ensure stability and minimal perturbation of backbone information.

Finally, $O_F,T_F$ pass through light projection layers $P_1,P_2$ and then into the shared UP:
\begin{equation}
\begin{aligned}
O_{UP} &= P_\psi\big(\mathrm{Fuse}_v(P_1(O_F), m_{\text{img}})\big), \\
T_{UP} &= P_\psi\big(\mathrm{Fuse}_t(P_2(T_F), m_{\text{text}})\big).
\end{aligned}
\label{eq:fuse}
\end{equation}

\paragraph{Why UniProj-Detv2 complements UniProj-Det.} The cross-fusion stage explicitly models early cross-modal interactions and can increase alignment capacity while remaining parameter-efficient: the cross-attention uses small projections and few heads (i.e., low FLOPs) and the shared UP leverages the fused features. In contrast to full joint fine-tuning, UniProj-Detv2 confines trainable parameters to a small, controlled set.

\subsection{Applying UniProj variants to an existing OVD (MDETR example)}
\begin{figure*}[!h]
    \centering
    \includegraphics[width=0.87\linewidth]{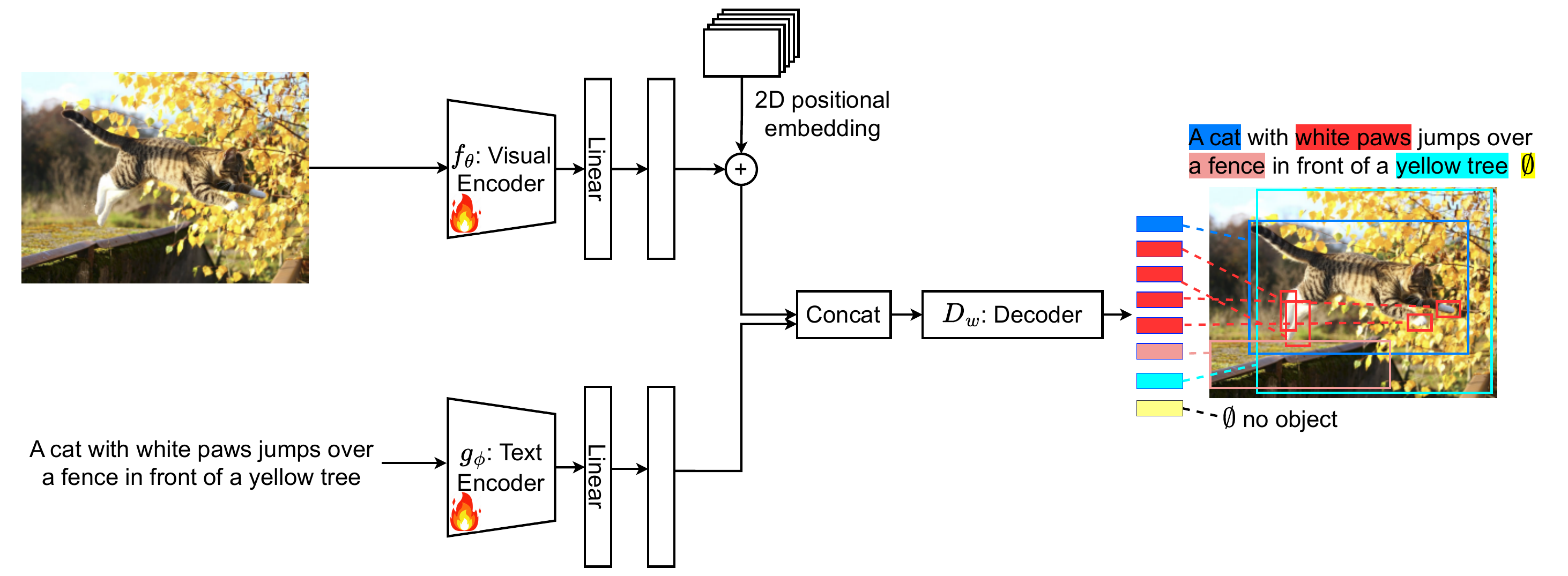}
    \vspace{0.25em}
    \includegraphics[width=0.87\linewidth]{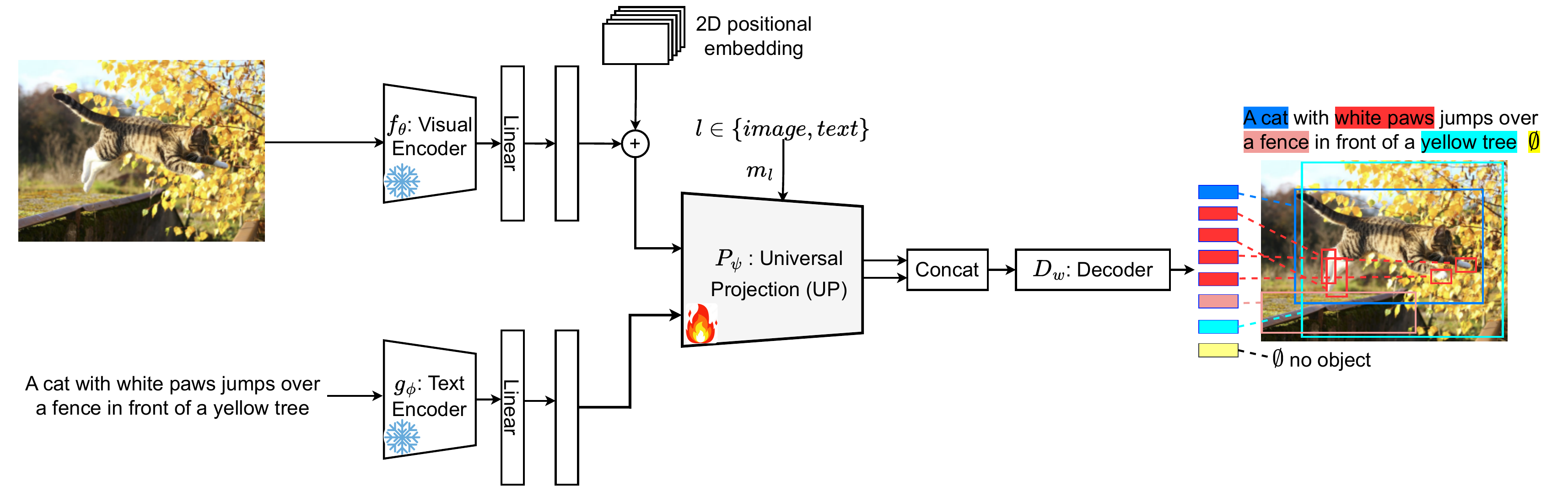}
    \vspace{0.25em}
    \includegraphics[width=1.14\linewidth]{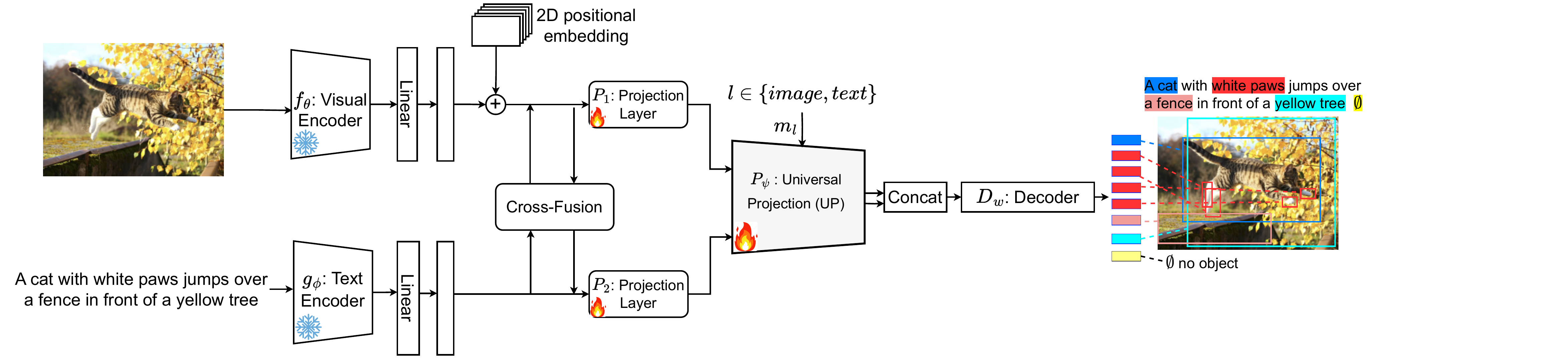}
    \caption{Top: (a) MDETR (full fine-tuning). Middle: (b) UniProj-Det (freeze backbones, train UP + modality tokens). Bottom: (c) UniProj-Detv2 (adds lightweight cross-fusion before UP).}
    \label{fig:mdetr-uniproj}
\end{figure*}

UniProj-Det is architecture-agnostic. To illustrate, Fig.~\ref{fig:mdetr-uniproj} shows three variants applied to MDETR: (a) original MDETR (full fine-tuning), (b) UniProj-Det applied to MDETR where the ResNet and RoBERTa are frozen and UP sits between encoders and the joint encoder, and (c) UniProj-Detv2 where a lightweight cross-fusion module precedes UP. In practice, the detector head (DETR decoder) can remain trainable or be frozen depending on the budget; our experiments report both settings.

\paragraph{Implementation notes.} In our implementation: UP is a 4-layer transformer encoder with reduced hidden dimension and 4 attention heads; modality tokens are vectors of dimension $d_m=64$; cross-fusion uses 2 attention heads with small projections. These design choices balance capacity and efficiency — ablations on depth/width are given in Section~\ref{sec:experiments}.

\paragraph{Summary.} UniProj-Det and UniProj-Detv2 offer a principled route to parameter-efficient OVD by (i) freezing large pretrained encoders, (ii) learning a small shared projection that maps both modalities to a task-specific space, and (iii) optionally adding a lightweight, low-cost cross-fusion stage to enhance cross-modal conditioning. In the next section we describe datasets, training details, and present empirical FLOPs/memory measurements and ablation studies that validate these claims.

\section{Experiment}
\label{sec:experiments}

In this section, we empirically validate the effectiveness of our proposed method UniProj-Det and its variant UniProj-Detv2, introduced in Section~\ref{sec:method}. We provide systematic comparisons with both open-vocabulary detectors (MDETR~\cite{kamath2021mdetr}, GLIP~\cite{li2022glip}) and efficient open-vocabulary detectors (Open-FGHA~\cite{shi2023edadet}, Open-Det~\cite{gu2021open}). To ensure fairness, all models employ a ResNet-101~\cite{he2016deep} as the vision encoder $f_\theta$ and a RoBERTa-base~\cite{liu2019roberta} as the text encoder $g_\phi$.\newline

\noindent\textbf{Training setup.}  
For MDETR and GLIP, both visual and textual backbones are trainable. Open-FGHA freezes the encoders but introduces multiple parameter-efficient modules: HiH-LoRA (3 blocks), Global Feature Enhancer (6 blocks), Bi-SCA (LayerNorm + Cross-Attention), and CQG dynamic anchors. Open-Det trains the ResNet backbone and several additional modules, including a LoRA head, Box Head, Binary Head, Transformer encoder-decoder, VLD-M, and a Generative Language Model.\newline
In contrast, UniProj-Det and UniProj-Detv2 freeze both encoders entirely and only update a lightweight Universal Projection (UP) module with 4 Transformer blocks. The DETR decoder $D_w$ remains frozen across all our variants. UniProj-Detv2 introduces a cross-modality fusion module prior to UP, as described in Section~\ref{sec:method}. For a fair comparison of computational efficiency, MDETR and GLIP are trained on 8 GPUs, whereas Open-FGHA, Open-Det, UniProj-Det, and UniProj-Detv2 are trained on 4 GPUs.  

\subsection{Pre-training Dataset}
Following MDETR~\cite{kamath2021mdetr}, all models are pre-trained on a large-scale combination of Flickr30k~\cite{young2014image}, MS COCO~\cite{lin2014microsoft}, and Visual Genome~\cite{krishna2017visual}. This provides rich region-level grounding annotations with millions of text–region pairs, ensuring consistent supervision across baselines and our methods.  

\subsection{Downstream Tasks}
We evaluate all methods on three downstream tasks: (1) phrase grounding, (2) referring expression comprehension, and (3) referring expression segmentation.  

\subsubsection{Phrase Grounding}
Phrase grounding requires mapping textual phrases to corresponding regions in an image. We evaluate on the Flickr30k Entities dataset, reporting Recall@$k$ (R@1, R@5, R@10). For each sentence, 100 candidate bounding boxes are predicted, ranked using soft token alignment scores. This allows us to measure how accurately the model grounds fine-grained linguistic tokens to visual regions.\newline
Table~\ref{tab:phrase_grounding_new} reports Recall@$k$ on Flickr30k Entities. Both UniProj-Det and UniProj-Detv2 match or surpass MDETR and GLIP while using significantly fewer trainable parameters. Compared to efficient methods Open-FGHA and Open-Det, our models consistently deliver higher accuracy with a simpler and lighter parameterization.  

\begin{table}[!h]
\centering
\resizebox{0.5\textwidth}{!}{
\begin{tabular}{lccc|ccc}
\hline
Method & \multicolumn{3}{c|}{Val} & \multicolumn{3}{c}{Test} \\
       & R@1 & R@5 & R@10 & R@1 & R@5 & R@10 \\
\hline
MDETR        & 82.5 & 92.9 & \textbf{94.9} & 83.4 & 93.5 & \textbf{95.3} \\
GLIP         & 83.1 & 93.2 & 94.5 & 83.8 & 93.6 & 95.0 \\
Open-FGHA    & 82.8 & 92.7 & 94.2 & 83.1 & 93.3 & 94.8 \\
Open-Det     & 83.0 & 92.9 & 94.3 & 83.6 & 93.4 & 94.9 \\
\textbf{UniProj-Det}    & 83.9 & 93.2 & 94.2 & \textbf{83.9} & 94.1 & 95.2 \\
\textbf{UniProj-Detv2}  & \textbf{84.0} & \textbf{93.6} & 94.9 & 83.8 & \textbf{94.6} & 95.2 \\
\hline
\end{tabular}}
\caption{Phrase grounding results on Flickr30k Entities. UniProj-Det and UniProj-Detv2 outperform efficient OVDs while matching the best OVD baselines.}
\label{tab:phrase_grounding_new}
\end{table}
\begin{figure*}[!h]
    \centering
    \includegraphics[width=0.8\linewidth]{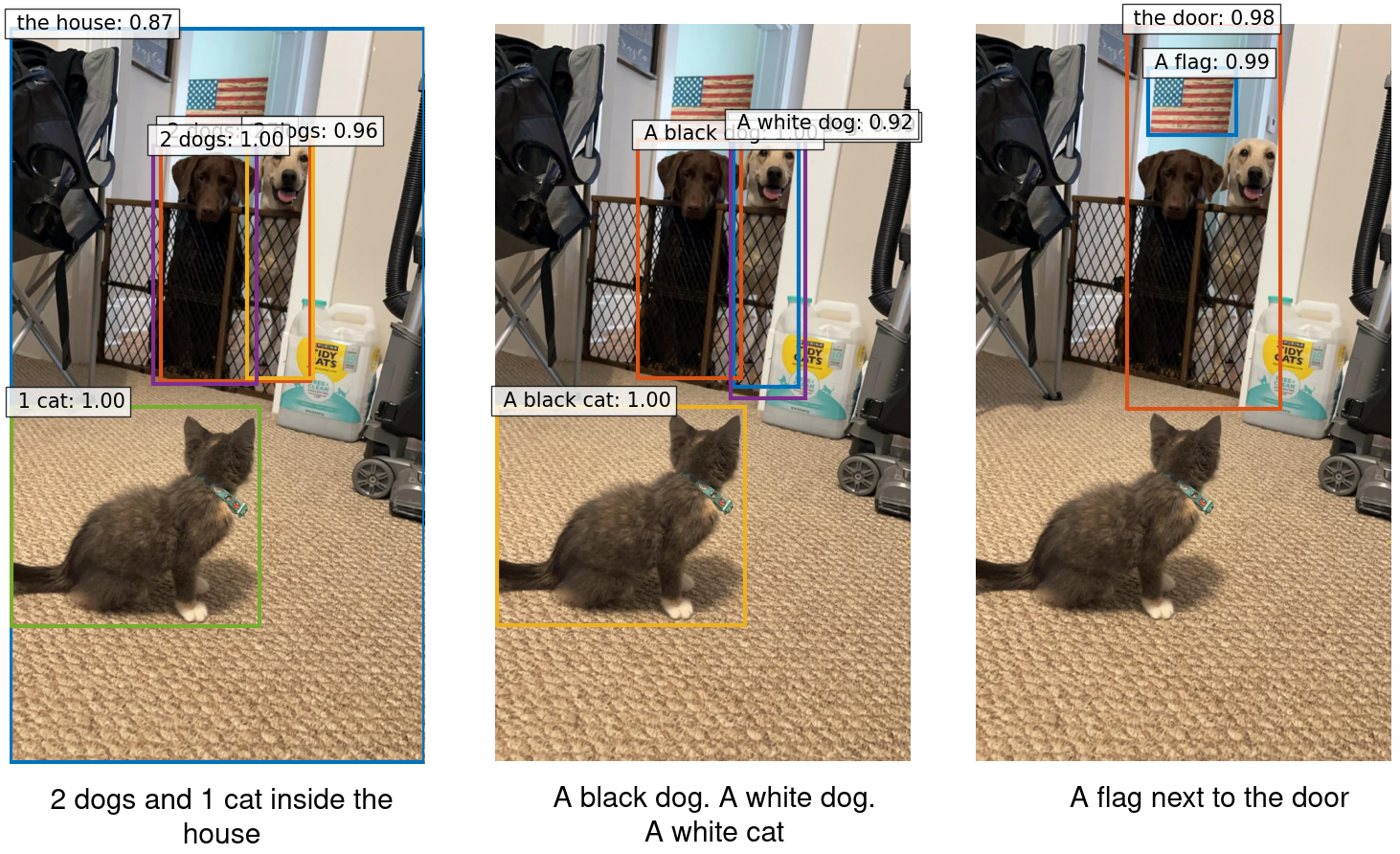}
    \caption{Illustration of phrase grounding using UniProj-Det. The model detects fine-grained regions corresponding to textual phrases, with positive token spans highlighted in green. This demonstrates the effectiveness of the universal projection and cross-modal alignment in mapping language to visual regions.}
    \label{fig:phrase_grounding}
\end{figure*}
Figure~\ref{fig:phrase_grounding} illustrates how UniProj-Det identifies the root of a phrase as the positive token span, highlighting its ability to accurately ground fine-grained textual references to image regions.

\subsubsection{Referring Expression Comprehension}
This task involves localizing a target object given a natural language expression. We use RefCOCO~\cite{kazemzadeh2014referit}, RefCOCO+~\cite{yu2016modeling}, and RefCOCOg~\cite{mao2016generation} datasets, following standard splits (person vs. object for testA/testB). Models predict 100 candidate boxes per image, ranked using the $\emptyset$ token for irrelevant boxes.\newline
Table~\ref{tab:ref_exp_new} compares performance on RefCOCO/RefCOCO+/RefCOCOg. Both UniProj-Det and UniProj-Detv2 reach competitive accuracy, demonstrating the benefit of the modality token and universal projection design. UniProj-Detv2 consistently outperforms efficient OVD baselines, highlighting the role of early cross-modal fusion.  

\begin{table}[!h]
\centering
\resizebox{0.5\textwidth}{!}{
\begin{tabular}{lccc|ccc|cc}
\hline
Method & \multicolumn{3}{c|}{RefCOCO} & \multicolumn{3}{c|}{RefCOCO+} & \multicolumn{2}{c}{RefCOCOg} \\
       & val & testA & testB & val & testA & testB & val & test \\
\hline
MAttNet~\cite{yu2018mattnet}    & 76.65 & 81.14 & 69.99 & 65.33 & 71.62 & 56.02 & 66.58 & 67.27 \\
ViLBERT~\cite{lu2019vilbert}    & -     & -     & -     & 72.34 & 78.52 & 62.61 & -     & -     \\
VL-BERT~\cite{su2020vl-bert}     & -     & -     & -     & 72.59 & 78.57 & 62.30 & -     & -     \\
UNITER~\cite{chen2020uniter}     & 81.41 & 87.04 & 74.17 & 75.90 & 81.45 & 66.70 & 74.86 & 75.77 \\
VILLA~\cite{gan2020large}      & 82.39 & 87.48 & 74.84 & 76.17 & 81.54 & 66.84 & 76.18 & 76.71 \\
ERNIE-ViL~\cite{yu2021ernie}   & -     & -     & -     & 75.95 & 82.07 & 66.88 & -     & -     \\
MDETR        & 86.8 & \textbf{89.6} & 81.4 & 79.5 & 84.1 & 70.6 & 81.6 & \textbf{80.9} \\
GLIP         & 86.7 & 89.0 & 81.5 & 79.3 & 83.7 & 70.5 & 81.4 & 80.5 \\
Open-FGHA    & 86.2 & 88.8 & 81.0 & 78.7 & 83.2 & 70.1 & 81.0 & 80.2 \\
Open-Det     & 86.4 & 88.9 & 81.2 & 78.9 & 83.5 & 70.4 & 81.2 & 80.4 \\
\textbf{UniProj-Det}   & 86.8 & 88.5 & \textbf{82.0} & \textbf{79.6} & 83.3 & 70.6 & \textbf{82.0} & 79.7 \\
\textbf{UniProj-Detv2} & \textbf{86.8} & 88.8 & 81.8 & 79.1 & \textbf{84.1} & \textbf{71.1} & 81.1 & 80.8 \\
\hline
\end{tabular}}
\caption{Accuracy performance comparison between our proposed models, UniProj-Det and UniProj-Detv2, and other detection models in the referring expression comprehension task on the RefCOCO, RefCOCO+, and RefCOCOg datasets. For testing, RefCOCO and RefCOCO+ datasets are evaluated using person vs. object splits: "testA" includes images with multiple people, while "testB" includes images with multiple objects from other categories. RefCOCOg features two distinct data partitions.}
\label{tab:ref_exp_new}
\end{table}

\subsubsection{Referring Expression Segmentation}
The goal is to segment objects described by textual expressions. We use the PhraseCut dataset, which contains images with multiple referring expressions and pixel-level masks. Training proceeds in two phases: (1) fine-tuning the bounding box predictor while keeping the backbone frozen, and (2) training a segmentation head supervised by Dice/F1 + Focal loss. During inference, masks are selected based on confidence scores and merged per expression.\newline
On PhraseCut, UniProj-Detv2 achieves the best balance of precision and mIoU, showing the advantage of cross-modal fusion for pixel-level localization (Table~\ref{tab:ref_exp_seg_new}).  

\begin{table}[!h]
\centering
\resizebox{0.5\textwidth}{!}{
\begin{tabular}{l|cccc}
\hline
Method & mIoU & Pr@0.5 & Pr@0.7 & Pr@0.9 \\
\hline
RMI~\cite{chen2019see} & 21.1 & 22.0 & 11.6 & 1.5 \\
HULANet~\cite{yu2018mattnet} & 41.3 & 42.4 & 27.0 & 5.7 \\
MDETR        & 53.1 & 56.1 & 38.9 & \textbf{11.9} \\
GLIP         & 52.8 & 55.7 & 38.5 & 11.7 \\
Open-FGHA    & 52.9 & 55.9 & 38.7 & 11.6 \\
Open-Det     & 53.0 & 56.0 & 38.8 & 11.7 \\
\textbf{UniProj-Det}   & 53.5 & 57.0 & 39.1 & 11.6 \\
\textbf{UniProj-Detv2} & \textbf{53.9} & \textbf{57.1} & \textbf{39.3} & 11.8 \\
\hline
\end{tabular}}
\caption{Validation of Referring Expression Segmentation using the mean intersection-over-union (IoU) between predicted and ground-truth masks, alongside precision Pr@$I$, where success is defined as the predicted mask achieving an IoU with the ground-truth that exceeds the threshold $I$.}
\label{tab:ref_exp_seg_new}
\end{table}
\begin{figure*}[!h]
    \centering
    \includegraphics[width=0.8\linewidth]{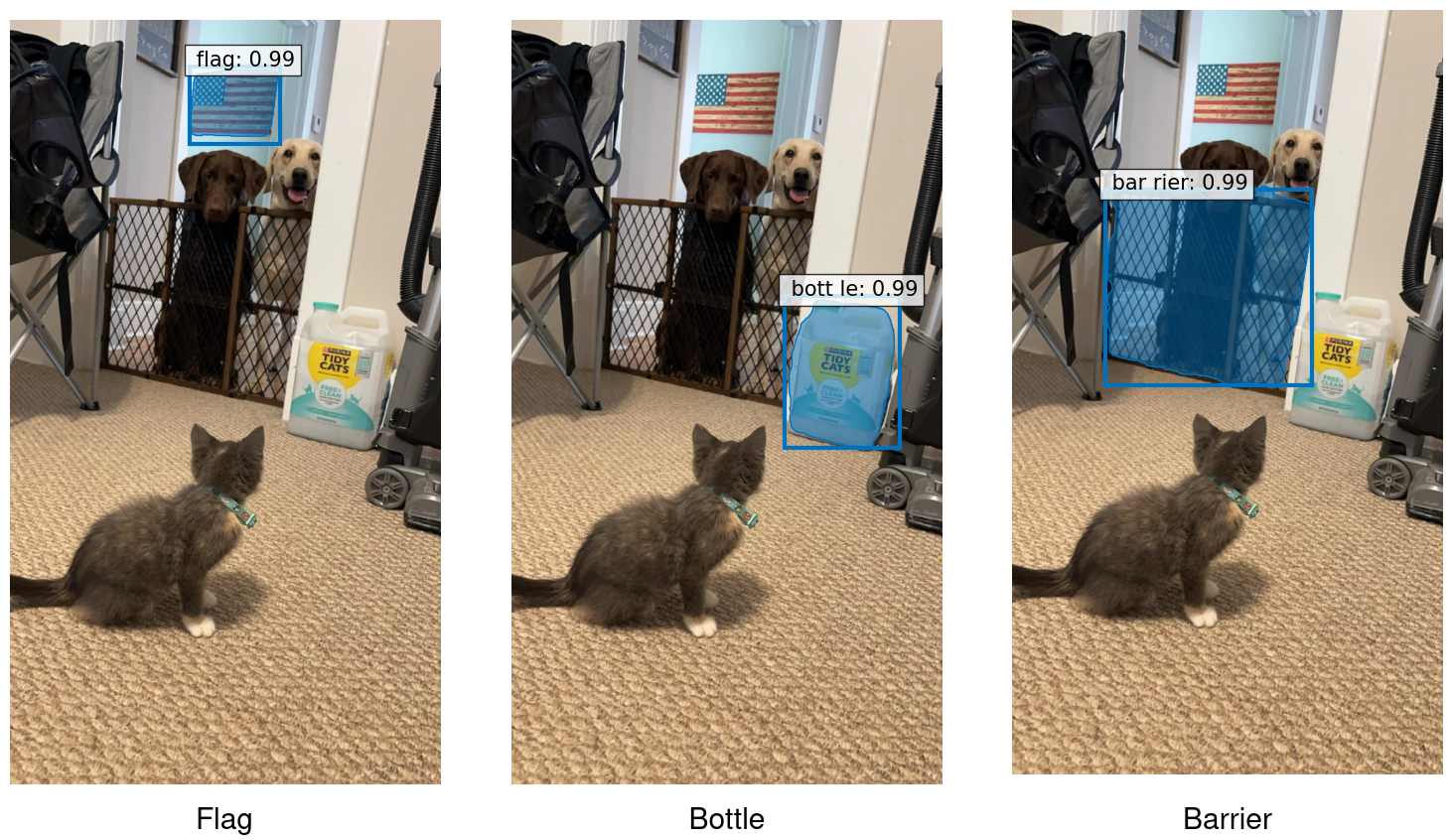}
    \caption{Referring expression segmentation examples with UniProj-Detv2. Predicted masks are overlaid on images for each referring expression. This highlights the model's ability to precisely localize and segment objects described by natural language, benefiting from the cross-modal universal projection and lightweight Transformer module.}
    \label{fig:segmentation}
\end{figure*}
Figure~\ref{fig:segmentation} visualizes segmentation outputs for referring expressions. Predicted masks are overlaid on images, showing how UniProj-Det accurately delineates objects corresponding to natural language descriptions.

\subsection{Efficiency Analysis}
\begin{figure*}[!h]
    \centering
    \includegraphics[width=0.75\linewidth]{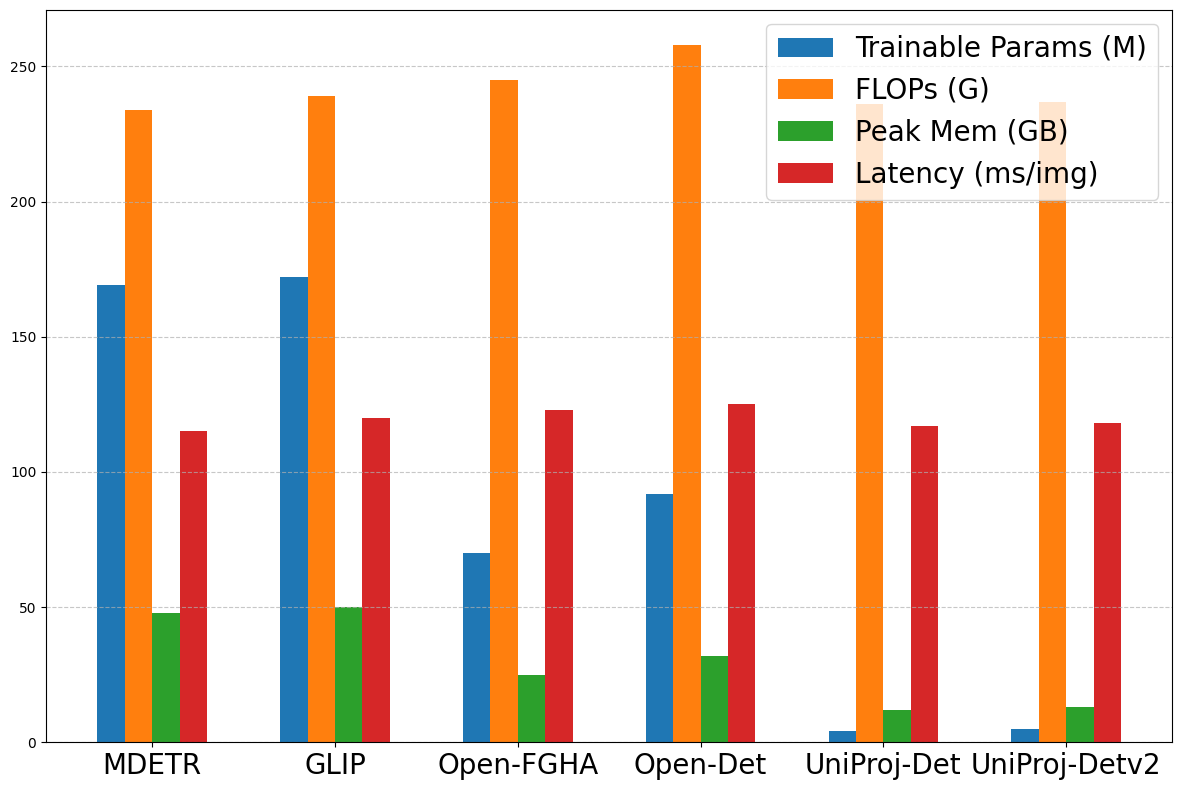}
    \caption{Efficiency comparison across methods. Metrics are reported as: millions of trainable parameters (M), gigaFLOPs per image (G), peak GPU memory in gigabytes (GB), and inference latency in milliseconds per image (ms/img). UniProj-Det and UniProj-Detv2 achieve significant reductions in trainable parameters and memory, with only marginal increases in FLOPs and latency.}
    \label{fig:efficiency}
\end{figure*}
We report computational efficiency in terms of trainable parameters, FLOPs per image, peak GPU memory during training, and inference latency. FLOPs denote the number of floating-point operations per forward pass, while trainable parameters (M) are measured in millions. Peak GPU memory is reported in gigabytes (GB), and inference latency is measured in milliseconds per image (ms/img). 

As shown in Figure~\ref{fig:efficiency}, UniProj-Det and UniProj-Detv2 reduce the number of trainable parameters by more than an order of magnitude and significantly lower training memory, while incurring only marginal overhead in FLOPs and latency. This demonstrates that our universal projection design achieves an optimal trade-off between efficiency and accuracy.

\section{Ablation Studies}
\label{sec:ablation}

In this section, we perform extensive ablations to analyze the design choices of UniProj-Det and UniProj-Detv2. All experiments are conducted on the Phrase Grounding task using the Flickr30k dataset. The evaluation metric is Recall@k as defined in the validation set (val).

\subsection{Effect of the Number of UP Layers}
The first ablation investigates how the depth of the UP module affects performance. We vary the number of Transformer blocks while keeping the backbones frozen and using addition fusion. The goal is to determine whether increasing UP depth improves the model’s capacity to adapt frozen backbone features for open-vocabulary grounding.

\begin{table}[!h]
\centering
\resizebox{0.5\textwidth}{!}{
\begin{tabular}{lccc|ccc}
\hline
UP Layers & \multicolumn{3}{c|}{UniProj-Det} & \multicolumn{3}{c}{UniProj-Detv2} \\
           & R@1 & R@5 & R@10 & R@1 & R@5 & R@10 \\
\hline
2  & 82.7 & 92.0 & 93.5 & 82.8 & 92.3 & 93.8 \\
4  & 83.98 & 93.15 & 94.20 & 84.02 & 93.56 & 94.9 \\
6  & 84.3 & 93.5 & 94.5 & 84.4 & 93.8 & 95.0 \\
8  & 84.5 & 93.7 & 94.7 & 84.6 & 94.0 & 95.1 \\
10 & 84.6 & 93.8 & 94.8 & 84.7 & 94.1 & 95.1 \\
\hline
\end{tabular}}
\caption{Ablation on the number of Transformer layers in the UP module. Increasing layers improves performance, but gains saturate after 8 layers.}
\label{tab:ablation_up_layers}
\end{table}

\textbf{Analysis:} Using only 2 UP layers underperforms significantly, indicating insufficient capacity to adapt frozen backbone features. The standard 4-layer UP already recovers strong performance, while increasing to 6–10 layers gives minor improvements. This suggests that a moderately deep UP (4 layers) is sufficient for effective phrase grounding, balancing performance and efficiency.  

\subsection{Impact of the Modality Token}
Next, we evaluate the effect of the modality token $m$, used to differentiate image and text features in the UP (see Equation~\ref{eq:fuse}). We remove the modality token while keeping the UP 4 layers, frozen ResNet and RoBERTa.

\begin{table}[!h]
\centering
\resizebox{0.5\textwidth}{!}{
\begin{tabular}{lccc|ccc}
\hline
Setting & \multicolumn{3}{c|}{UniProj-Det} & \multicolumn{3}{c}{UniProj-Detv2} \\
        & R@1 & R@5 & R@10 & R@1 & R@5 & R@10 \\
\hline
With modality token &  83.98 & 93.15 & 94.20 & 84.02 & 93.56 & 94.9 \\
Without modality token & 65.0 & 68.7 & 63.2 & 64.23 & 65.45 & 65.98 \\
\hline
\end{tabular}}
\caption{Effect of the modality token. Removing it significantly degrades performance because the UP cannot properly distinguish between visual and textual features.}
\label{tab:ablation_modality_token}
\end{table}

\textbf{Analysis:} The modality token allows the UP to condition its projections on the feature type (image vs text). Removing it leads to a large drop ($\sim 15$\% R@1), showing that this mechanism is critical for cross-modal adaptation.

\subsection{Fusion Method in UP}
We investigate different fusion strategies between image and text features: addition, concatenation, and cross-attention. All models use 4 UP layers and frozen backbones (ResNet and RoBERTa).

\begin{table}[!h]
\centering
\resizebox{0.5\textwidth}{!}{
\begin{tabular}{lccc|ccc}
\hline
Fusion & \multicolumn{3}{c|}{UniProj-Det} & \multicolumn{3}{c}{UniProj-Detv2} \\
       & R@1 & R@5 & R@10 & R@1 & R@5 & R@10 \\
\hline
Addition &  83.98 & 93.15 & 94.20 & 84.02 & 93.56 & 94.9 \\
Concatenation & 83.5 & 92.8 & 93.9 & 83.6 & 93.0 & 94.2 \\
Cross-Attention & 83.8 & 93.67 & 94.26 & 83.9 & 93.3 & 94.92 \\
\hline
\end{tabular}}
\caption{Ablation of fusion methods in UP. Addition is simple yet effective, concatenation slightly underperforms, cross-attention slightly improves over concatenation but adds computational cost.}
\label{tab:ablation_fusion}
\end{table}

\textbf{Analysis:} Concatenation requires additional projection layers to maintain alignment and slightly decreases performance. Cross-attention improves over concatenation but introduces higher computational cost. Addition offers a strong trade-off between simplicity, efficiency, and accuracy, justifying its use in the main experiments.

\subsection{Backbone Choice}
Finally, we study the impact of different backbone pairs on phrase grounding. We fix UP to 4 layers with addition fusion and modality tokens.

\begin{table}[!h]
\centering
\resizebox{0.5\textwidth}{!}{
\begin{tabular}{lccc|ccc}
\hline
Backbone (Vision, Text) & \multicolumn{3}{c|}{UniProj-Det} & \multicolumn{3}{c}{UniProj-Detv2} \\
                         & R@1 & R@5 & R@10 & R@1 & R@5 & R@10 \\
\hline
ResNet, RoBERTa &  83.98 & 93.15 & 94.20 & 84.02 & 93.56 & 94.9 \\
ResNet, BERT    & 83.2 & 92.7 & 93.8 & 83.3 & 92.9 & 94.0 \\
Swin, RoBERTa   & 84.2 & 93.5 & 94.6 & 84.3 & 93.8 & 95.0 \\
Swin, BERT      & 83.8 & 93.0 & 94.3 & 83.9 & 93.3 & 94.6 \\
\hline
\end{tabular}}
\caption{Ablation on different backbone pairs. Powerful backbones slightly improve performance, but a smaller backbone with a well-designed UP is sufficient.}
\label{tab:ablation_backbone}
\end{table}

\textbf{Analysis:} While Swin improves performance slightly, the frozen ResNet with RoBERTa already yields strong results, demonstrating that the UP module effectively compensates for smaller backbone capacity. This confirms that a moderate backbone is sufficient for open-vocabulary grounding when combined with a trainable UP.

\subsection{Conclusion of Ablations}
Overall, these ablations show that:
\begin{itemize}
    \item 4-layer UP is a strong, efficient choice; increasing depth gives diminishing returns.
    \item The modality token is essential for distinguishing image vs text features.
    \item Simple addition fusion is sufficient, with more complex fusions offering minor gains at higher cost.
    \item Moderate backbone capacity is adequate; UP can adapt frozen features effectively.
\end{itemize}
These findings justify our main experimental settings for UniProj-Det and UniProj-Detv2.

\section{Conclusion}
\label{sec:conclusion}

We introduced UniProj-Det and its enhanced variant UniProj-Detv2, lightweight and parameter-efficient frameworks for open-vocabulary object detection. By freezing vision and language backbones and training only a compact Universal Projection (UP) module, our models achieve performance comparable to fully-trained OVD baselines (MDETR, GLIP) while outperforming existing efficient OVD methods (Open-FGHA, Open-Det).\newline

Ablation studies show that a 4-layer UP with addition fusion and modality tokens is sufficient, and further depth or complex fusion offers minimal gains. Our results indicate that strong backbone representations are beneficial but not strictly necessary, as the UP module effectively adapts moderate features for open-vocabulary detection.\newline

Future work includes scaling to larger multimodal datasets, exploring adaptive UP architectures, and applying UniProj-Det in real-time or resource-constrained scenarios.

\section*{Acknowledgments}
We would like to express our deepest gratitude to \textbf{LabCom IRISER} for their financial support, which made this work possible as part of an internship. Their commitment to fostering research and innovation has been invaluable throughout this project. We are extremely thankful for the opportunity and the resources they provided, enabling us to pursue and complete this work successfully.

\bibliographystyle{ieeetr}
\bibliography{bibliography}

\clearpage

\end{document}